\documentclass[10pt,twocolumn,letterpaper]{article}

\usepackage{iccv}
\usepackage{times}
\usepackage{epsfig}
\usepackage{graphicx}
\usepackage{amsmath}
\usepackage{amssymb}
\usepackage{array}
\usepackage{pdfpages}
\usepackage{booktabs}
\usepackage{adjustbox}
\usepackage{caption}
\usepackage[symbol]{footmisc}

\usepackage{xcolor}
\usepackage{footnote}

\usepackage[breaklinks=true,bookmarks=false]{hyperref}

\iccvfinalcopy % *** Uncomment this line for the final submission

\def\iccvPaperID{****} % *** Enter the ICCV Paper ID here

\ificcvfinal\fi
\begin{document}
% \begin{savenotes}

%%%%%%%%% TITLEhttps://www.overleaf.com/project/6154431a9bdc8177a16771ef
\title{ViSeRet: A simple yet effective approach to moment retrieval via fine-grained video segmentation}

\author{Aiden Seungjoon Lee\\
Twelve Labs\\
Seattle, U.S.A.\\
{\tt\small aiden@twelvelabs.io}
% For a paper whose authors are all at the same institution,
% omit the following lines up until the closing ``}''.
% Additional authors and addresses can be added with ``\and'',
% just like the second author.
% To save space, use either the email address or home page, not both
\and
Hanseok Oh\\
KAIST\\
Seoul, Republic Of Korea\\
{\tt\small hanseok@kaist.ac.kr}
\and
Minjoon Seo\\
KAIST\\
Seoul, Republic Of Korea\\
{\tt\small minjoon@kaist.ac.kr}
}

\maketitle
% Remove page # from the first page of camera-ready.
\ificcvfinal\thispagestyle{empty}\fi

%%%%%%%%% ABSTRACT
\begin{abstract}
\textit{Video-text retrieval has many real-world applications such as media analytics, surveillance and robotics. This paper presents the 1st place solution to the video retrieval track of the ICCV VALUE Challenge 2021. We present a simple yet effective approach to jointly tackle two video-text retrieval tasks (video retrieval and video corpus moment retrieval) by leveraging the model trained only on the video retrieval task. In addition, we create an ensemble model that achieves the new state-of-the-art performance on all four datasets (TVr, How2r, YouCook2r, and VATEXr) presented in the VALUE Challenge.}
\end{abstract}

%%%%%%%%% BODY TEXT
\section{Introduction}
The video retrieval track of the ICCV VALUE Challenge 2021 \cite{VALUE} tackles two types of text-to-video tasks: video retrieval (VR) and video corpus moment retrieval (VCMR). The goal of both tasks is to find the best video that describes the given query(caption) sentence within an extensive video database. For the VCMR task, the model should additionally output the start and end time within the video, making the task more challenging. 

The competition is comprised of four datasets: TVr, How2r, YouCook2r, and VATEXr \cite{Tvr, HERO, Yc2r, Vatex}. TVr and How2r datasets are designed as VCMR tasks, while YouCook2r and VATEXr datasets are designed as VR tasks. The metric for both tasks is the average value of Recall$@$1, Recall$@$5, and Recall$@$10 (i.e., average recall). For the VCMR task, the retrieved video is considered correct only if the temporal intersection over union (tIoU) between the ground truth (GT) timestamp and the predicted timestamp is greater than 0.7 (tIoU $>$ 0.7). The overall rank of the model is determined by averaging the rank of the model’s performance on each dataset (i.e., average rank). Our contributions to the competition are the following: 
\begin{itemize}
    \item We propose a simple yet effective approach, ViSeRet (Video Segment Retrieval), to tackle video moment localization retrieval (VCMR) task using the model trained only for the video retrieval (VR) task. We find that our video retrieval model alone with a simple heuristic segmenting technique can outperform the previous state-of-the-art model optimized for the VCMR task. 
    \item An ensemble of ViSeRet and HERO \cite{HERO} achieves the new state-of-the-art performance on all four datasets in the VALUE Challenge Retrieval Track.
\end{itemize}

%------------------------------------------------------------------------
% trim={1cm 7cm 1cm .8cm}
\begin{figure*}[h]
    \centering
\resizebox{2\columnwidth}{!}{%
    \includegraphics[width=15cm,height=5cm,trim={1cm 3cm 1cm 4.5cm},clip]{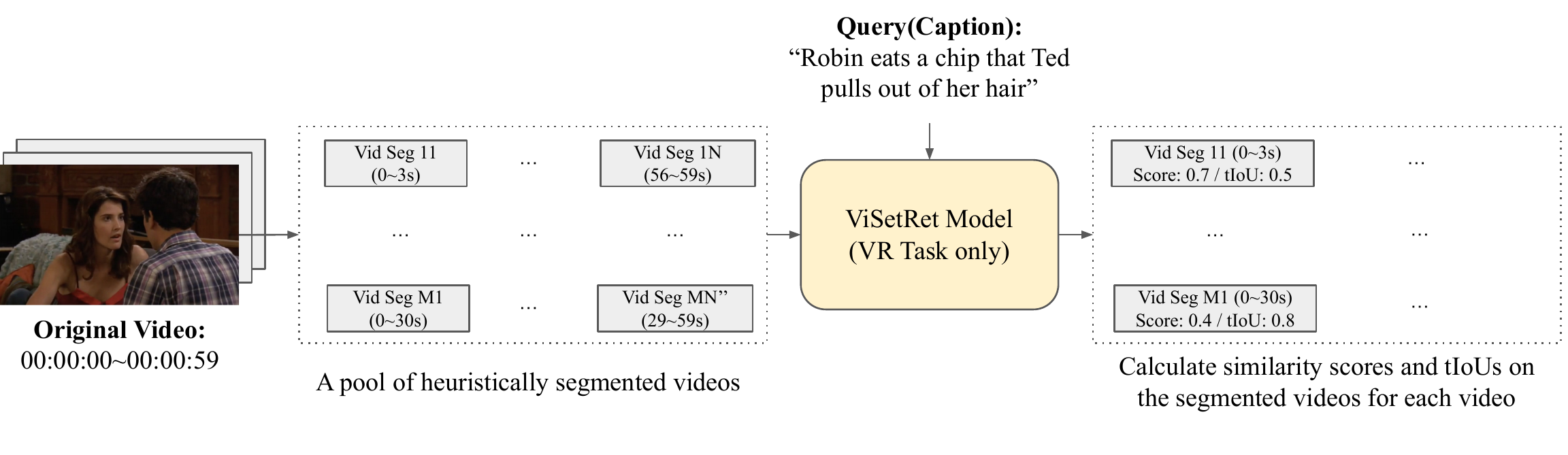}
    }
    \caption{Our approach for tackling VCMR task using the model trained for VR task.}
    \label{approach-vcmr}
\end{figure*}

\section{ViSeRet Model}
Figure \ref{approach-vcmr} presents our approach. Our approach is to reformat the video corpus moment retrieval(VCMR) task into the traditional video retrieval (VR) task, and use the model trained for VR task to jointly solve the two tasks. Our assumption is that the model maximized for the VR task should ideally perform competitively in the VCMR task with a simple heuristic video segmentation technique. 

To be specific, we segment each video into several fixed-length (i.e., 5 seconds, 10 seconds) video clips, treat them as separate videos for our video retrieval model to rank them, and use the pre-segmented timestamp information to calculate the tIoU with the ground truth(GT) timestamp. For the VR task, we simply use the entire video segment for the same video retrieval model to rank them. We find that our video-retrieval-only approach outperforms the previous state-of-the-art models trained on the VCMR task not only on the VR task datasets (YouCook2r and VATEXr), but also in the VCMR task datasets (TVr and How2r).

%-------------------------------------------------------------------------
\subsection{Model architecture}
\begin{figure*}[h]
    \centering
\resizebox{2\columnwidth}{!}{%
    \includegraphics[width=15cm,height=8cm,trim={0 0 0 1cm},clip]{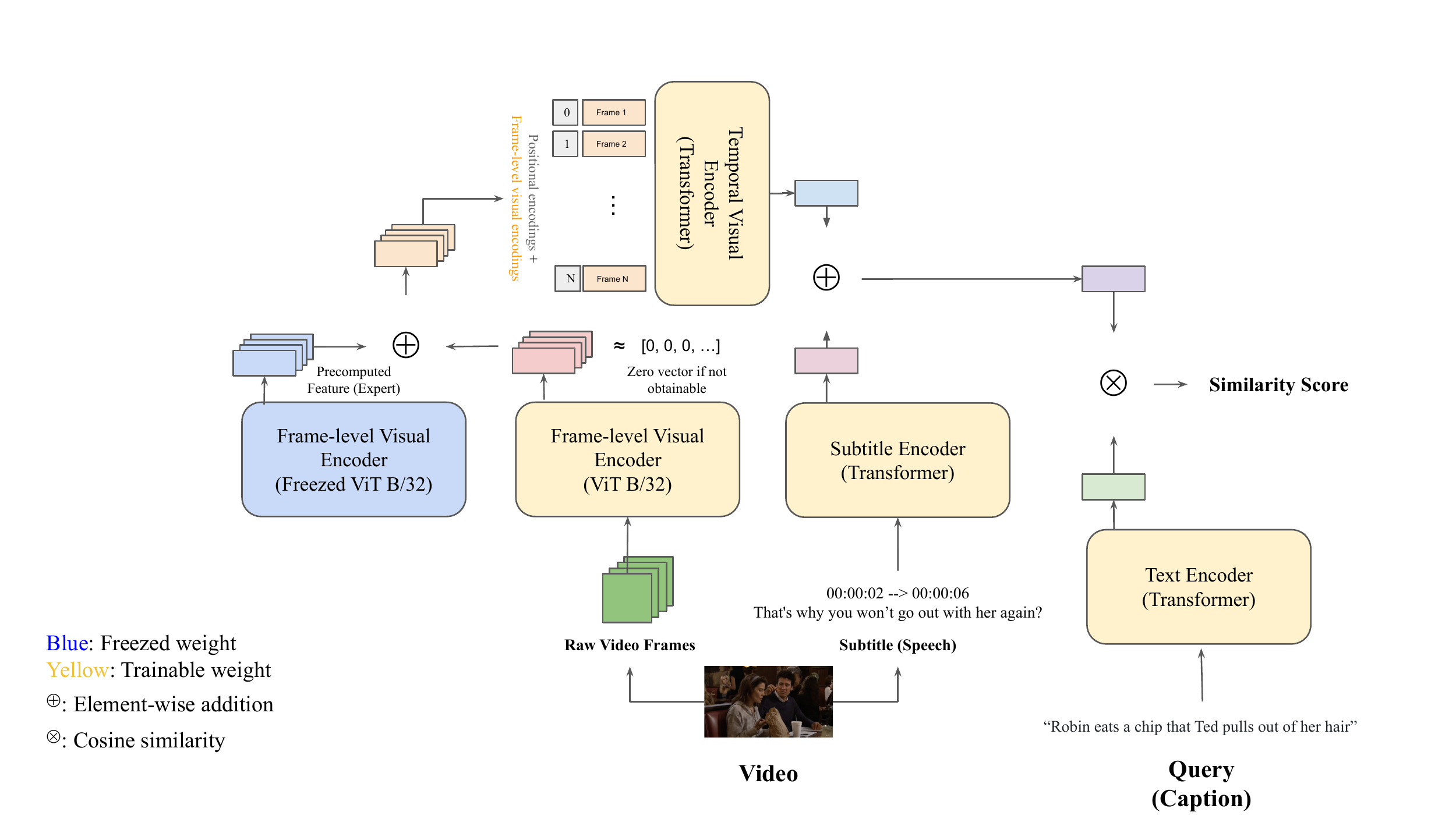}
    }
    \caption{Our ViSeRet model architecture.}
    \label{viseret-model}
\end{figure*}

The model architecture of ViSeRet is illustrated in Figure \ref{viseret-model}. Our model is inspired by the recent success of transferring the knowledge from a large-scale image-text pre-trained model, such as CLIP \cite{CLIP}, to video retrieval task \cite{MDMMT, CLIP4CLIP}. 

\textbf{Visual Encoding} For each frame, we obtain two types of CLIP-based features: precomputed CLIP feature and raw frame CLIP feature that is directly extracted from the raw video. The difference between the two features is whether the weight of the CLIP model is updated or not during the training. The precomputed CLIP feature is obtained from a separate expert-like freezed CLIP model while the raw frame CLIP feature is obtained through the CLIP model where its weight is updated during the training process. We elementwise add these two CLIP features and use it as the final frame-level visual features. There are two reasons for using two types of CLIP features. First, the elementwise-sum structure of the precomputed CLIP feature and the raw frame CLIP feature creates a residual-connection-like structure for the CLIP model. Second, we can robustly handle missing data where the original raw videos are not obtainable due to copyright and privacy issues by replacing each raw frame CLIP feature with a zero vector.

After the visual encodings are obtained for each frame, we create temporal connections among them using Transformer and obtain a single video-level visual encoding vector.

\textbf{Subtitle Encoding} We obtain a single BERT-based \cite{BERT} subtitle encoding vector([CLS] token) for the entire video.

\textbf{Visual-Subtitle Fusion Encoding} We elementwise add the visual encoding vector with the subtitle encoding vector and use this vector as the final video encoding. We find this simple approach enough to see a noticeable performance difference. We leave it as future work to evaluate other fusion methods \cite{MMT, HERO}.

\textbf{Similarity Estimation} We use CLIP's text encoder to obtain the query (caption) embedding vector. The similarity is calculated using cosine similarity between the query (caption) embedding with the final video embedding. 

\subsection{Training} We train our model based on the segmented video-query(caption) pairs for VCMR task datasets TVr and How2r, and the original video-query(caption) pairs for YouCook2r and VATEXr(VR task) datasets. The videos are segmented based on their ground truth (GT) start and end time steps for each query (caption) annotation.

\begin{table}
\begin{center}
% \begin{adjustbox}{width=1\columnwidth,height=0.5cm}
\resizebox{\columnwidth}{!}
% \resizebox{0.6\textwidth}{!}
{%
\begin{tabular}{ lcc}
\hline
& TVr & How2r\\
\hline
Lengths(s) & 3$/$5$/$10$/$20$/$30$/$60 & 5$/$10$/$20$/$30 \\
Strides(s) & 1$/$2$/$3$/$5$/$8$/$10 & 2$/$5$/$10$/$15\\
$\#$ of videos& 2.2k & 1k\\
$\#$ of video segments & 330k & 48k \\
\% of annotations covered & 86\% & 82\%\\
\hline
\end{tabular}
}
% \end{adjustbox}
\caption{The lengths and strides we used for the TVr and How2r dataset to segment the videos. Each video is segmented into the lengths and strides specified in the table. For example, a 60 seconds long video in the How2r dataset is segmented into [0$\sim$5s, 2$\sim$7s, 4$\sim$9s,...] video clips. }
\label{segment-info}
\end{center}
\end{table}

\subsection{Inference} During the inference time, the model is given a query (caption) and a pool of videos without any specific timestamp information. Therefore, for VCMR task datasets, we segment each video into several fixed-length video clips (i.e., 3s, 5s, 10s, ...) and give these segmented videos as input to the model. The specific video clip lengths are determined based on the timestamp distribution of the validation data. We also make sure that the new pool of segmented videos  include at least 80\% of the annotation's ground truth timestamp-video pairs that satisfy tIoU $>$ 0.7. The specific numbers that we used for the TVr and How2r datasets are presented in Table \ref{segment-info}.

This new pool of segmented videos allows the model trained for the VR task to perform VCMR tasks by using the video segment's timestamp information to calculate the tIoU with the ground truth timestamp.

\section{Experiments}

\subsection{Experiment Setups}
The model is trained on each dataset(TVr, How2r, VATEXr, and YouCook2r) independently using noisy contrastive estimation(NCE) loss function. The batch size is 64, with the learning rate set to 1e-4. We set the maximum frame length and subtitle token length to 64 and 512 and truncated the rest. We use the pre-trained BERT-base model for subtitle encoding, CLIP ViT-B/32 model for visual encoding, and CLIP text model for query(caption) encoding. 

%-------------------------------------------------------------------------

\begin{table}
\begin{center}
\resizebox{\columnwidth}{!}{%
\begin{tabular}{ lcccc}
% \begin{tabular}{c|>{\centering\arraybackslash}p{2cm}|>{\centering\arraybackslash}p{2cm}|>{\centering\arraybackslash}p{2cm}|>{\centering\arraybackslash}p{2cm}|>{\centering\arraybackslash}p{2cm}}
\hline
 & TVr & How2r & YouCook2r & VATEXr \\
\hline
HERO (FT only)&8.81&2.13&42.37&30.22\\
Ours (FT only)&9.77&7.74&50.92&55.46\\
\hline
Relative Gain&\textbf{+10.8\%}&\textbf{+263.3\%}&\textbf{+20.1\%}&\textbf{+83.5\%}\\
\hline
\end{tabular}
}
\end{center}
\caption{Our ViSeRet finetune only(FT only) model outperforms the baseline  HERO finetune only(FT only) model on all four datasets by a large margin.. The units are measured in average recall (higher the better).}
\label{finetune-result}
\end{table}

\subsection{Single Model Results}
Table \ref{finetune-result} compares our finetune only(FT only) model with the baseline HERO model on the VALUE datasets. For a fair comparison, we compare our model with the finetune only (FT only) model. Our model exceeds the baseline model performance on all four datasets with up to 263.3\% relative performance gain in the How2r dataset.

Table \ref{submission-result} compares the performance of our submitted single models with the previous state-of-the-art, HERO (AT$\rightarrow$ST, PT$\rightarrow$FT). Our single model achieves better performance on three out of the four VALUE datasets (How2r, YouCook2r, VATEXr). Due to the time constraint, we only had the chance to pretrain our model on the subset of the HowTo100m dataset \cite{HowTo100m, CUPID} and finetune it on the YouCook2r dataset. We leave it as future work to perform large-scale pretraining on other external datasets and finetune the model to each dataset.

\subsection{Ensemble Model Results}
\textbf{Methodology} Our ensemble model combines ViSeRet (single) and HERO(AT$\rightarrow$ST, PT$\rightarrow$FT. For each dataset, we find the best weight coefficient combination $\alpha$ and $\beta$ and multiply them to each model's prediction (logit) scores for each output. Then, we merge the outputs of the two models and re-rank them based on the weighted scores. The best weight coefficient $\alpha$ and $\beta$ are found through a grid search on the validation data.

\textbf{Results} Table \ref{submission-result} presents the result of our submitted ensemble model. Our ensemble model achieves 22.9\% performance increase on average compared to our single model, achieving the new state-of-the-art across all four datasets. In addition, our ensemble model ranks the first place on How2r and VATEXr dataset, the second place on YouCook2r dataset, and the third place on TVr dataset, winning the video retrieval track for VALUE Challenge 2021 with the mean rank of 1.75.  

%-------------------------------------------------------------------------

\begin{savenotes}
\begin{table}
\begin{center}
\resizebox{\columnwidth}{!}{%
\begin{tabular}{ lcccc}
% \begin{tabular}{c|>{\centering\arraybackslash}p{2cm}|>{\centering\arraybackslash}p{2cm}|>{\centering\arraybackslash}p{2cm}|>{\centering\arraybackslash}p{2cm}|>{\centering\arraybackslash}p{2cm}}
\hline
 & TVr & How2r & YouCook2r & VATEXr\\
\hline
% HERO(AT$\rightarrow$ST, PT$\rightarrow$FT)&13.56&3.95&54.28&49.09\\
% Ours (single)&9.77&7.74&55.73$^\ast$&55.46\\
% Ours (ensemble)&14.18&7.74$^\ast$$^\ast$&62.72&58.03\\
HERO(AT$\rightarrow$ST, PT$\rightarrow$FT)&13.56&3.95&54.28&49.09\\
Ours (single)&9.77&\textbf{7.74}&55.73 \footnote[1]{The difference between our model results in Table 1 for the YouCook2 dataset is due to additional pretraining on the subset(25k videos) of the HowTo100m dataset.}&55.46\\
Ours (ensemble)&\textbf{14.18}&\textbf{7.74}\footnote[7]{Due to the time constraint, we didn't have enough time to obtain the ensemble model's result on How2r dataset. Instead, we submitted the same value from our single model result.}&\textbf{62.72}&\textbf{58.03}\\
\hline
\end{tabular}
}
\end{center}
\caption{Our final submission results to the VALUE Challenge 2021 Leaderboard. The units are measured in average recall (higher the better). Our ViSeRet-ensemble model outperforms the previous state-of-the-art(SOTA) on all four VALUE datasets.}
\label{submission-result}
\end{table}
\end{savenotes}

% \lfoot{The difference between our model results in Table 1 for the YouCook2 dataset is due to additional pretraining on the subset(25k videos) of the HowTo100m dataset.}
% \lfoot{Due to the time constraint, we didn't have enough time to obtain the ensemble model's result on How2r dataset and instead submitted the single model result.}

\section{Conclusion and Future Work}
In this report, we show that our video retrieval model ViSeRet can perform effectively on video moment localization retrieval (VCMR) tasks with a simple heuristic segmenting technique. Our finetune only(FT only) model shows a significant performance increase on all four datasets of VALUE Challenge compared to the baseline HERO fine-tune only model with up to 263\% relative increase in How2r dataset. Our ensemble model that is composed of ViSeRet and HERO achieves the new state-of-the-art on all four datasets, winning the video retrieval track of ICCV Value Challenge 2021.

Future work includes a large-scale pretraining of ViSeRet model on other external datasets. It is also worth noting that our model might benefit from incorporating a separate reader model that extracts moments from the results of ViSeRet. Due to the time constraint, we did not have enough time to incorporate the reader model into the final submission, but we briefly describe what we tried here. 

Using the same embedding vectors of query and video as that of the video retriever model, the reader model can easily compute query-video similarity scores. On top of the scores we can use classifier layers, which can be convolution 1D filters following HERO model's reader setting \cite{HERO} to modify the retrieval model's results. Because we found that using a retriever model with segmented clips could result in better VCMR performance than baseline, we set a threshold on the re-ranking score to choose whether to use the output of the reader model. When re-rank score is above the threshold, which can be interpreted as the video reader model being reliable, we use the reader-predicted time span to extract a subset of the retriever's outputs. Otherwise, we use time segment information of the segmented video clips as the model output instead. We expect that the reader model trained end-to-end with the retriever can improve the end-to-end performance over the current proposed model.

{\small
\bibliographystyle{ieee_fullname}
\bibliography{main}
}
% \end{savenotes}
\end{document}